\documentclass[runningheads]{llncs}
\usepackage{graphicx}
\usepackage{amssymb}
\usepackage{pifont}
\usepackage{amsmath}
\usepackage{todonotes}
\usepackage{hyperref}
\usepackage[misc]{ifsym}

\begin{document}
\title{Neural Fields for 3D Tracking of Anatomy and Surgical Instruments in Monocular Laparoscopic Video Clips}
\titlerunning{3D Anatomy and Surgical Instrument Tracks Using Neural Fields}
%
\author{Beerend G.A. Gerats\inst{1,2} $^\text{(\Letter)}$ \and
Jelmer M. Wolterink\inst{3,4} \and
Seb P. Mol\inst{1, 4} \and
Ivo A.M.J. Broeders\inst{1,2}}


\authorrunning{B.G.A. Gerats et al.}
\institute{AI \& Data Science Center, Meander Medical Center, The Netherlands \email{bga.gerats@meandermc.nl} \and
Robotics and Mechatronics, University of Twente, The Netherlands \and
Department of Applied Mathematics, University of Twente, The Netherlands \and
Technical Medical Center, University of Twente, The Netherlands}
\maketitle              
\begin{abstract} 
Laparoscopic video tracking primarily focuses on two target types: surgical instruments and anatomy. The former could be used for skill assessment, while the latter is necessary for the projection of virtual overlays. Where instrument and anatomy tracking have often been considered two separate problems, in this paper, we propose a method for joint tracking of all structures simultaneously. Based on a single 2D monocular video clip, we train a neural field to represent a continuous spatiotemporal scene, used to create 3D tracks of all surfaces visible in at least one frame. Due to the small size of instruments, they generally cover a small part of the image only, resulting in decreased tracking accuracy. Therefore, we propose enhanced class weighting to improve the instrument tracks. We evaluate tracking on video clips from laparoscopic cholecystectomies, where we find mean tracking accuracies of 92.4\% for anatomical structures and 87.4\% for instruments. Additionally, we assess the quality of depth maps obtained from the method's scene reconstructions. We show that these pseudo-depths have comparable quality to a state-of-the-art pre-trained depth estimator. On laparoscopic videos in the SCARED dataset, the method predicts depth with an MAE of 2.9 mm and a relative error of 9.2\%. These results show the feasibility of using neural fields for monocular 3D reconstruction of laparoscopic scenes. 
\end{abstract}

\begin{figure}[t]
    \centering
    \includegraphics[width=\textwidth]{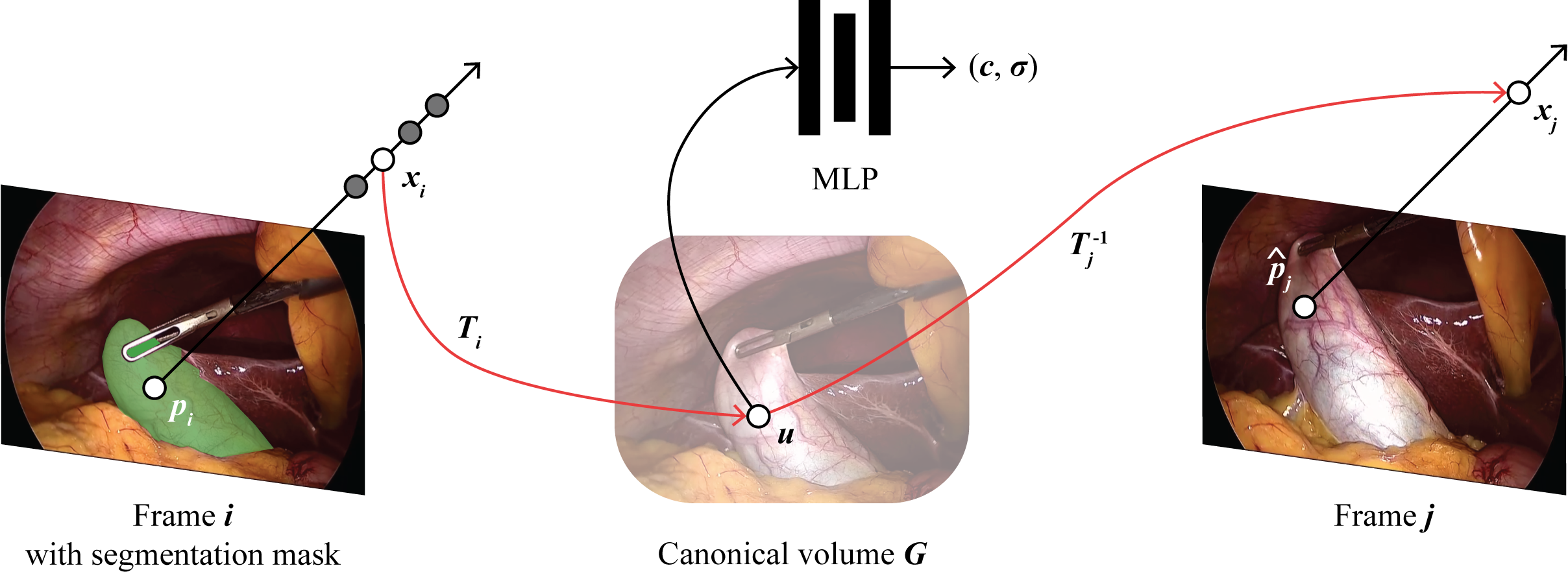}
    \caption{Overview of our method. A pixel $p_i$ is selected in frame $i$, from where a ray is cast through the virtual scene. A sampled location $x_i$ is mapped to location $u$ in canonical volume $G$. This location is used to predict color $c$ and material density $\sigma$. To predict the translation of this point to frame $j$, the location is mapped via an inverse transform and reprojected to the predicted pixel location $\hat{p}_j$.}
    \label{fig:method_overview}
\end{figure}

\section{Introduction}
Optical tracking of elements in laparoscopic videos provides valuable information for computer-assisted surgery. Tracking efforts primarily focus on two target types: surgical instruments and anatomy. E.g., the former is used for skill assessment  \cite{fathollahi2022video,lavanchy2021automation}, while the latter is necessary for the projection of virtual overlays \cite{pelanis2021evaluation}. 

Several 2D instrument trackers have been proposed~\cite{du2018articulated,kurmann2017simultaneous,robu2021towards}, but these trackers generally lack depth perception and fail to track points adequately in 3D. This might be overcome by the inclusion of markers~\cite{cartucho2022enhanced} or electromagnetic tracking~\cite{liu2016laparoscopic}. However, such approaches require adaptations to the instruments or surgical workflow and are not widely available. 3D pose trackers designed for correcting position errors, e.g. in robotic systems, require a known pose at the start of tracking~\cite{du2016combined}. Other methods use camera parameters to calculate 3D instrument locations~\cite{zhao2017tracking}, which are typically unknown for regular surgery. In anatomical tracking,  some approaches use an additional depth sensor~\cite{golse2021augmented}. However, these are difficult to implement in-vivo. Alternatively, depth could be estimated from stereoscopic \cite{long2021dssr} or monocular video~\cite{beilei2024surgical,shao2022self}. Although the depth estimates are highly informative, these methods do not model tissue deformation. Methods for 3D scene reconstruction based on simultaneous localization and mapping, do model for anatomical deformations~\cite{li2020super,long2021dssr,song2017dynamic}, but are limited in handling complex deformations and color changes.

While instrument and anatomy tracking have often been considered two separate problems, here, we propose a method for joint tracking of \textit{all} elements in a laparoscopic video. Based on a 2D video clip, we track the 3D location of all surfaces that are visible in at least one frame of the video. Key to our approach is the use of neural fields, which represent continuous spatio-(temporal) scenes in the weights of neural networks. Neural fields have previously been used in laparoscopic scene representation~\cite{wang2022neural,yang2023neural,zha2023endosurf}. A limitation of these prior works is the need for stereoscopic vision, which limits the applicability of these methods in standard minimally invasive procedures. Instead, we propose to leverage the recently proposed OmniMotion~\cite{wang2023omnimotion} approach, in which monocular videos are used to reconstruct a 3D + time scene representation, without the need for camera calibration, stereoscopic vision, or depth maps. We evaluate the tracking performance on video clips from laparoscopic cholecystectomies for various tissue and instrument types. We show that instrument tracks can be obtained using enhanced class weighting. Last, we assess the quality of the depth maps obtained from the method's scene reconstructions. We show that these pseudo-depths have comparable quality to a state-of-the-art pre-trained depth estimator.

\section{Methods}
We propose to use neural fields as representations of surgical scenes visualized in monocular laparoscopic video clips. The neural fields are optimized using OmniMotion~\cite{wang2023omnimotion}, with additional class weighting to improve surgical instrument tracking. An overview of the method is given in Figure~\ref{fig:method_overview}.

\subsection{Neural Fields for Pixel Tracking}\label{sec:omnimotion}
We use OmniMotion \cite{wang2023omnimotion} for the reconstruction of 3D + time scene representations from 2D videos. The method captures camera movement and scene deformations in monocular video by mapping through a canonical volume $G$. Camera rays are cast through the scene at frame $i$, along which 3D locations $x_i$ are sampled. Using bijective mapping $u = T_i(x_i)$, the 3D locations are translated to location $u$ in the canonical volume. The mappings are invertible such that the new position of this point in frame $j$ can be calculated with $x_j = T_j^{-1}(T_i(x_i))$. A canonical location $u$ is used to query a multi-layer perception (MLP) that returns material density~$\sigma$ and color~$c$ for this point. When all sampled points along a camera ray are mapped to canonical space and queried to the MLP to obtain their densities and colors, the quadrature rule is used to find the predicted color~$\hat{C}_i$ for the corresponding pixel $p_i$ \cite{mildenhall2021nerf}. The predicted flow of a pixel between two frames $\hat{f}_{i \rightarrow j} = \hat{p}_j - p_i$ can be calculated similarly, using the material density values of the sampled points along the ray. In this way, the model can account for occlusions. The predicted flows are supervised with flows from a pre-trained optical flow model (RAFT) \cite{teed2020raft} (Equation~\ref{equ:loss_flow}), while the predicted pixel colors are supervised with the training images (Equation~\ref{equ:loss_color}).

\begin{equation} \label{equ:loss_flow}
    L_{\text{flow}} = \sum_{f_{i \rightarrow j} \in \Omega_p} \| \hat{f}_{i \rightarrow j} - f_{i \rightarrow j} \|_1
\end{equation}

\begin{equation} \label{equ:loss_color}
    L_{\text{color}} = \sum_{(i, p) \in \Omega_p} \| \hat{C}_i(p) - C_i(p) \|_2^2
\end{equation}

Although the method is supervised with 2D flows, the method's representation of scene deformation is in 3D. This makes it possible to track locations on anatomical structures and to follow surgical instruments over the depth axis. Because of the flow regularization, the method does not require camera parameters, stereoscopic video, or pre-computed depth images. Additionally, as camera movement is compensated for implicitly by the method, camera locations are not needed as well. This makes the method applicable to video clips from a regular laparoscope. Note that for each scene, we optimize an individual model.

\subsection{Segmentation Masks}
Since OmniMotion simultaneously tracks all pixels in the video, it is not possible to distinguish instruments from anatomy pixels. Therefore, we propose to combine the method with segmentation masks that cover the elements of interest. The segmentation masks can be manually annotated or found by pre-trained segmentation models \cite{grammatikopoulou2024spatio}. Note that the masks are not necessary for training and can be applied afterward unless class weighting is used (see next section). Only a single mask at the start of the video is required to track these pixels throughout the rest of the video. An example mask can be seen in Figure~\ref{fig:method_overview}.

\subsection{Class Weighting}
One of our goals is to track surgical instruments through the video. However, due to their size, instruments generally cover a small part of the image only. In the regular training approach of neural radiance fields, there is no incentive to represent a pixel more accurately than others. This means that reconstructions of irrelevant background pixels are reconstructed with similar importance as the surgical instruments. Therefore, we propose to use weighting in the loss function, where the weight depends on the class of the pixel, i.e. ``instrument'' or ``background'':

\begin{equation}
    L = w_{\text{class}} ( L_{\text{flow}} + \lambda L_{\text{color}} ) + L_{\text{other}}
\end{equation}

where $w_{\text{class}}$ = 1 if the pixel is in the background and a larger value otherwise, $\lambda$ is a loss weighting factor, and $L_{\text{other}}$ is a combination of regularizing losses. The pixel classes could be obtained by manual or automatic segmentation masks.

\subsection{Data}
We evaluate our model on two tasks. First, we assess the method's performance in 2D tracking of anatomical structures and surgical instruments. We train OmniMotion on all 80-frame video clips in the CholecSeg8k dataset \cite{hong2020cholecseg8k}. The set includes 101 short videos of 25fps that display laparoscopic cholecystectomies. Each video frame is accompanied by manually annotated segmentation masks of tissues and instruments. We pre-process the videos by cropping to a rectangle, to remove the circular border of the scope as much as possible, and resizing to 256~$\times$~256 pixels. After pre-processing, we find 84 clips that contain a segmentation in the first frame for the ``gallbladder'' class. For ``liver'' and ``gastrointestinal tract`` we find 101 and 46 clips, respectively. For the instruments ``grasper'' and ``l-hook electrocautery'' there are 65 and 24 clips available, respectively. In total, we track 231 anatomical structures and 89 surgical instruments.

Second, we explore the accuracy of the third axis, by comparing predicted depths with ground-truth depth measurements. For the evaluation of the depth predictions, we use laparoscopic videos of the SCARED dataset \cite{allan2021stereo}. The videos display a static surgical scene, recorded by various moving cameras. We select the first camera of each scene, as we evaluate the method for the monocular setting. The videos in the dataset are attributed with ground-truth depth values, measured via a structured light pattern projected on the field of view. We train OmniMotion on all nine videos individually, where the video length ranges between 88 and 728 frames.

\subsection{Evaluation Metrics}
For 2D tracking, we calculate the mean accuracy of pixels in a segmentation mask that are tracked throughout the video. For each pixel $p_1$ in the segmentation mask of the first frame, we get a predicted pixel location $\hat{p}_j$ at consecutive frames $j \in [2, 3, ..., 80]$. Predictions that are located within the segmentation mask of that frame are considered accurate, while those that fall outside the mask are inaccurate. Predicted locations that are outside the image boundaries are excluded from evaluation. In this way, we calculate the per-frame accuracy. When no segmentation mask is available, e.g. when an instrument is temporarily out of view, the frame is discarded. Finally, we calculate the mean accuracy by averaging the per-frame accuracies over all frames first, and over all videos afterwards.

For the evaluation of the predicted depth maps, we follow the general approach for evaluating pseudo-depths \cite{ranftl2020towards}, meaning that we shift and scale the depth predictions towards the ground-truth values before calculating the metrics.

\begin{figure}[t]
\includegraphics[width=\textwidth]{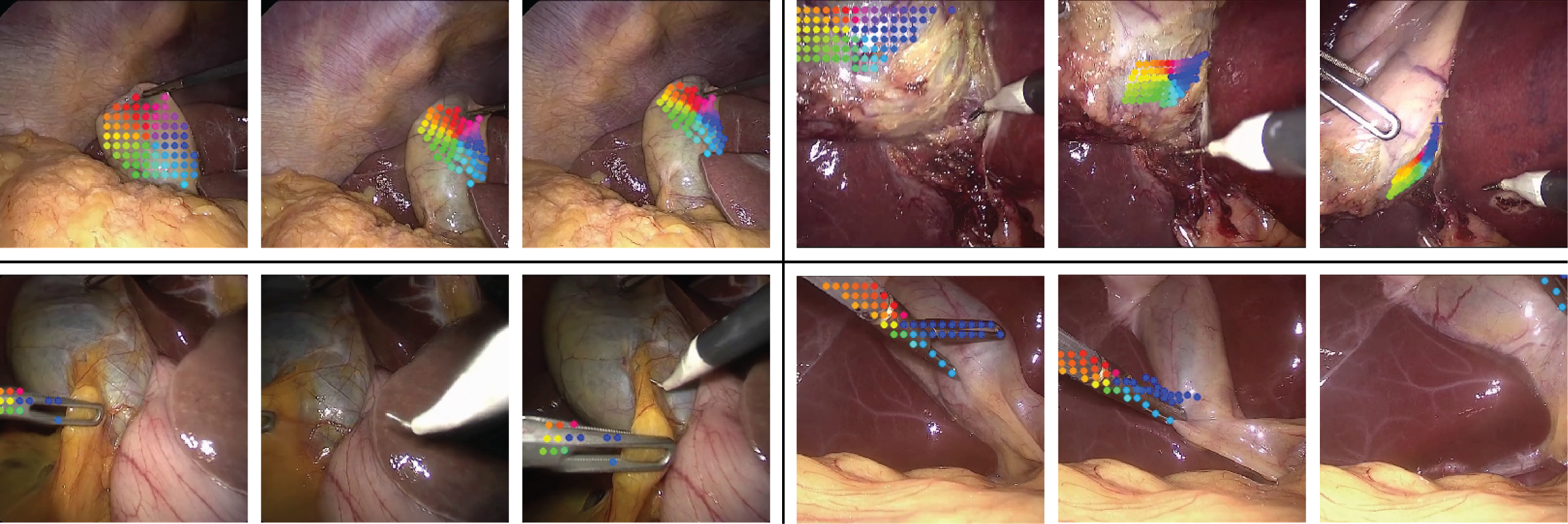}
\caption{Examples of tracked pixels (top: gallbladder, bottom: grasper) through 80-frame video clips, with the first, middle and last frame displayed here. The full videos are available via: \url{https://vimeo.com/920225544}.} \label{fig:tracking_qualitative}
\end{figure}

\section{Experiments and Results}
We train the method in batches of 256 correspondences from 8 image pairs. Along each camera ray, 32 points are sampled. Learning rates $3 \times 10^{-4}$ and $1 \times 10^{-4}$ are used for color and flow loss, respectively. Using an NVIDIA A40 GPU, it takes $\pm$10 minutes for RAFT pre-processing an 80-frame video clip. Training OmniMotion for 10K iterations requires an additional $\pm$45 minutes.

\begin{figure}[t]
\centering
\includegraphics[width=\textwidth]{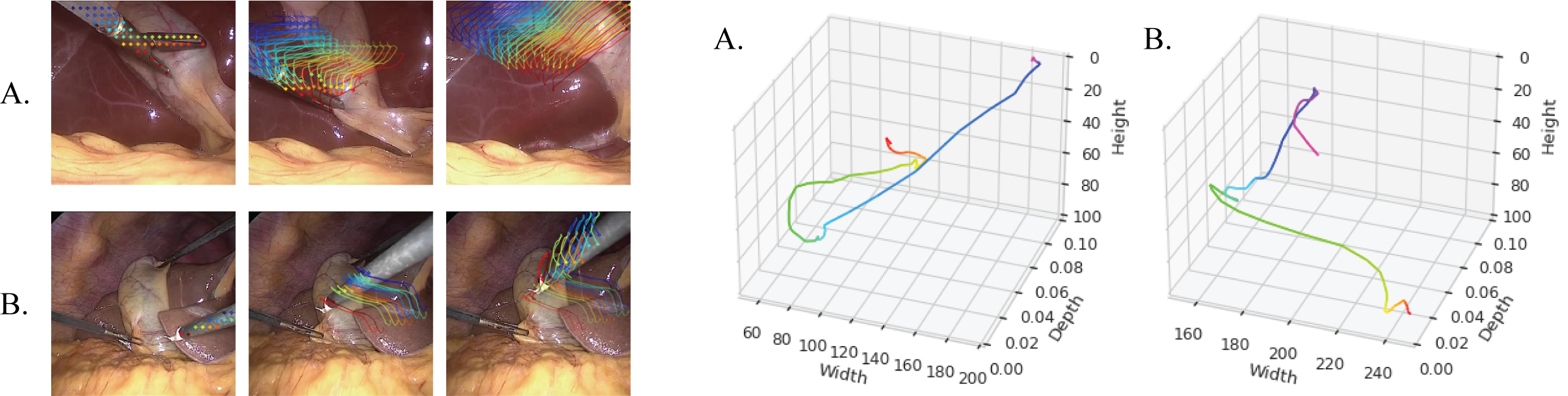}
\caption{Tracks of surgical instruments. Left: first, middle and last frame of the video, with trails following the instrument. Right: instrument tracks visualized in 3D.} \label{fig:3d_tracks}
\end{figure}

\subsection{Evaluation of Anatomy and Instrument Tracking}
A selection of qualitative results is given in Figure~\ref{fig:tracking_qualitative}. The top row displays two videos where gallbladder pixels are tracked. In the left example, the tracks accurately follow the gallbladder deformation under retraction with a grasper. In the right example, the camera zooms out while the gallbladder is released from the grasper grip. The pixel tracks accurately follow the anatomical structure. In the bottom row, the grasper is tracked. Left, the instrument leaves the field of view during the middle section of the video. It can be seen that the method accurately re-identifies the grasper when re-entering the field of view. In the right example, the grasper is heavily displaced to the upper-right corner in the last frame, where only a small part of the instrument remains available. Also in this scenario, the tracks follow the target pixels accurately. Instrument movements are visualized in Figure~\ref{fig:3d_tracks} for two example clips. In the 3D plots, we show the median location of the instrument over time. It can be seen that the method captures movements over three axes.

\begin{table}[b]
\centering
\caption{Mean accuracy (\%) of 2D tracking of anatomical structures and surgical instruments in CholecSeg8k. Instruments are tracked with different class weights $w_{\text{class}}$. Note that $w_{\text{class}}$=1 means that no class weighting is used.}
\label{tab:results_tracking}
\begin{tabular}{|l|c|c|c|c|} \hline
 & $w_\text{class}$=1 & $w_\text{class}$=5 & $w_\text{class}$=10 & $w_\text{class}$=20 \\ \hline
Gallbladder      & 91.7 ($\pm$ 11.8) & -           & -           & -           \\
Liver            & 93.6 ($\pm$ 8.7)  & -           & -           & -           \\
Gastroint. tract & 90.9 ($\pm$ 15.0) & -           & -           & -           \\
Grasper          & 83.6 ($\pm$ 23.0) & 84.3 ($\pm$ 21.3) & 85.1 ($\pm$ 19.8) & 85.9 ($\pm$ 20.0) \\
L-hook           & 83.6 ($\pm$ 28.8) & 92.0 ($\pm$ 21.7) & 92.5 ($\pm$ 20.6) & 91.5 ($\pm$ 20.3) \\ \hline
All anatomies    & 92.4 ($\pm$ 11.4) & -           & -           & -           \\
All instruments  & 83.6 ($\pm$ 24.7) & 86.3 ($\pm$ 21.7) & 87.1 ($\pm$ 20.3) & 87.4 ($\pm$ 20.3) \\ \hline
\end{tabular}
\end{table}

Table~\ref{tab:results_tracking} provides the quantitative measurements in tracking performance of anatomies and instruments. With a mean accuracy of 92.4\%, the method is able to track a large portion of the anatomy pixels accurately. Although most instrument pixels are properly tracked as well, the tracking performance is lower. The drop in performance can be explained by the small image area that the instruments typically cover. The effect can be counterbalanced with class weighting, where instrument tracking performance gradually increases from 83.6\% to 87.4\% for larger weights.

\begin{figure}[t]
\centering
\includegraphics[width=0.75\textwidth]{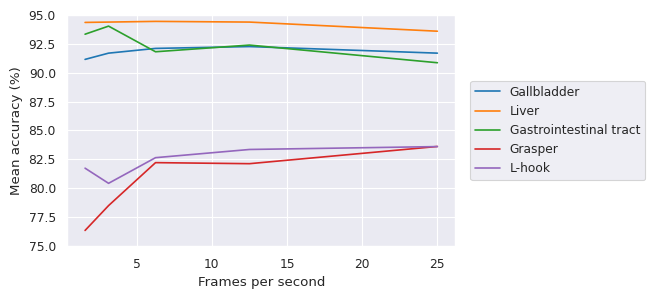}
\caption{Performance in 2D tracking for various temporal resolutions.} \label{fig:fps}
\end{figure}

Tracking accuracy for different frame rates is given in Figure~\ref{fig:fps}. Here, we decrease the temporal resolution of 25fps gradually to 50, 25, 12.5, and 6.25 percent of the original. Where the tracking performance for the anatomical structures remains largely unaffected, the quality of the instrument tracks decreases with fewer frames per second. This effect could be explained by the more drastic movements that instruments generally make in comparison to the anatomy.

\begin{table}[b]
\caption{Error of pseudo-depth estimation on surgical scenes in SCARED dataset, in comparison with a pre-trained dense depth estimator (DPT) \cite{ranftl2021vision}.}\label{tab:results_scared}
\begin{tabular}{|c|c|c|c|c|c|c|} \hline
    Scene & \multicolumn{2}{|c|}{MAE (mm) $\downarrow$} & \multicolumn{2}{|c|}{AbsRel (\%)$\downarrow$} & \multicolumn{2}{|c|}{$\delta < 1.25$ (\%)$\uparrow$} \\
    no. & OmniMotion & DPT & OmniMotion & DPT & OmniMotion & DPT \\ \hline
    1 & 2.9 & 2.2 & 7.8  & 6.2  & 95.5 & 99.0 \\
    2 & 1.4 & 1.5 & 4.8  & 4.9  & 97.8 & 99.8 \\
    3 & 5.3 & 5.0 & 14.6 & 13.0 & 77.0 & 85.4 \\
    4 & 1.7 & 2.0 & 7.4  & 8.1  & 97.1 & 96.3 \\
    5 & 2.1 & 4.2 & 7.8  & 17.3 & 95.5 & 77.8 \\
    6 & 2.5 & 2.2 & 7.2  & 6.4  & 96.2 & 98.8 \\
    7 & 5.1 & 4.5 & 13.9 & 13.0 & 80.0 & 84.6 \\
    8 & 2.2 & 2.1 & 7.5  & 7.3  & 96.7 & 96.0 \\
    9 & 2.7 & 1.6 & 12.0 & 6.8  & 90.3 & 96.5 \\ \hline
    Average & 2.9 ($\pm$ 1.3) & 2.8 ($\pm$ 1.3) & 9.2 ($\pm$ 3.2) & 9.2 ($\pm$ 4.0) & 91.8 ($\pm$ 7.4) & 92.7 ($\pm$ 7.5) \\ \hline
\end{tabular}
\end{table}

\subsection{Evaluation of Reconstructed Pseudo-Depth}

\begin{figure}[t]
\centering
\includegraphics[width=0.75\textwidth]{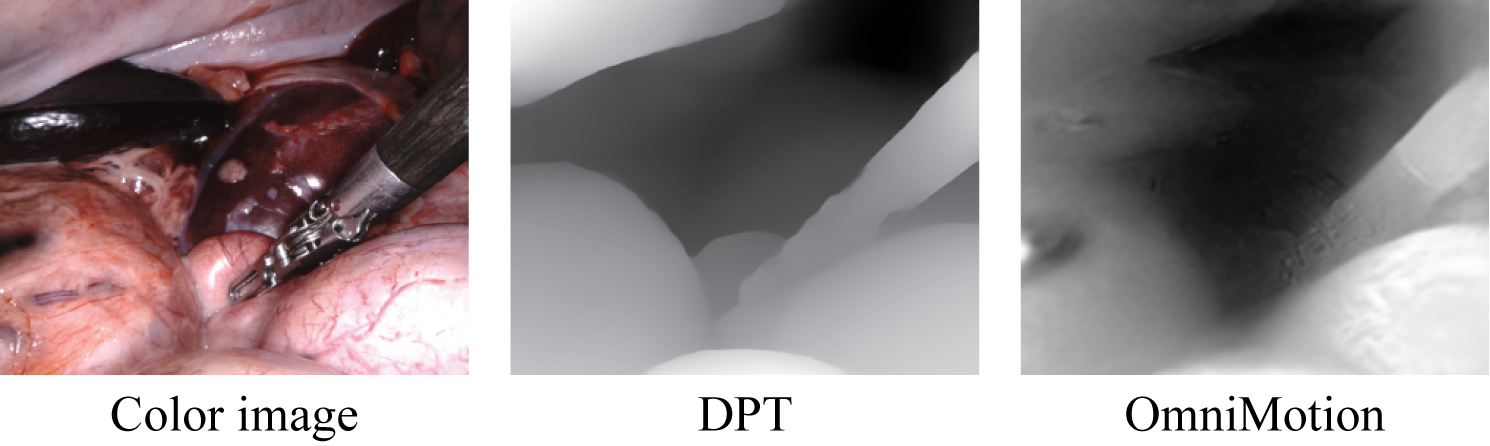}
\caption{An example frame from the SCARED dataset, with disparity estimated by DPT \cite{ranftl2021vision} and reconstructed with OmniMotion.} \label{fig:depth_qualitative}
\end{figure}

To assess the quality of the third pixel-tracking axis, we compare the estimated depth values with ground-truth measurements. Table~\ref{tab:results_scared} shows the error in depth estimation for all nine videos in the SCARED dataset \cite{allan2021stereo}. On average, the estimates have a mean average error of 2.9 mm and a relative error of 9.2\% to the actual depth. This performance is comparable to depth estimates by DPT \cite{ranftl2021vision}, a state-of-the-art pre-trained transformer for monocular depth estimation. The comparison shows that the depth axis of the pixel tracks is relatively accurate as well. Figure~\ref{fig:depth_qualitative} shows an example frame, with depth estimates by DPT and OmniMotion. Although the method's depth values are less smooth, they represent an accurate geometry of the underlying scene.

\section{Discussion \& Conclusion}
We combined tissue deformation modeling and surgical instrument tracking by tracking all pixels in a laparoscopic video at once. We used neural fields for 3D scene reconstruction, which could be trained without the need for camera calibration, camera locations, stereoscopic vision, or depth maps. In contrast to previously proposed methods for 3D scene reconstructions, this makes the method applicable as-is to standard minimally invasive procedures. In our experiments, we show that the combination of OmniMotion \cite{wang2023omnimotion} and segmentation masks can be used for constructing accurate 3D anatomy and instrument tracks. The quality of the instrument tracks can be improved by using class weighting, where instrument pixels are weighted more heavily during optimization.

A limitation of the used method is the relatively slow training speed. Ideally, the scene reconstructions are built in real-time, such that the resulting tracks are available during surgery. Yang et al. \cite{yang2023neural} have shown that drastic speed-ups, over a 100$\times$ faster, are possible by leveraging multi-plane fields. Moreover, the use of Gaussian splatting \cite{kerbl20233d} could potentially result in even faster training and real-time rendering.

A direction for future work is a deeper exploration of uses for neural fields from surgical videos. For example, when scene reconstructions are accurate enough, anatomical structures in pre-operative scans could potentially be mapped to the neural representation. Because neural fields can model tissue deformation, virtual overlays could be rendered that follow the underlying structures. Another example is the use of 3D anatomical tracks for the calculation of stress applied to the tissue, which could be visualized or used for haptic feedback. In our work, we have shown the feasibility of using neural fields for 3D monocular scene reconstructions, which could form the basis for these further explorations.

\section*{Declarations}
Research at the Meander AI \& Data Science Center was sponsored by Johnson \& Johnson MedTech. Jelmer M. Wolterink was supported by NWO domain Applied and Engineering Sciences VENI grant (18192).


%
%
%
%
\bibliographystyle{splncs04}
\bibliography{bibliography}

\end{document}